\documentclass[namedreferences]{kluwer}
\newcommand{\ignore}[1]{}

\input{psfig.sty}

\begin{document}
\begin{article}
\begin{opening}
\title{Tagger Evaluation Given Hierarchical Tag Sets}
\author{I. Dan \surname{Melamed}\email{dan.melamed@westgroup.com}}
\institute{West Group}
\author{Philip \surname{Resnik}\email{resnik@umiacs.umd.edu}}
\institute{University of Maryland}
\begin{abstract}
We present methods for evaluating human and automatic taggers that
extend current practice in three ways.  First, we show how to evaluate
taggers that assign multiple tags to each test instance, even if they
do not assign probabilities.  Second, we show how to accommodate a
common property of manually constructed ``gold standards'' that are
typically used for objective evaluation, namely that there is often
more than one correct answer.  Third, we show how to measure
performance when the set of possible tags is tree-structured in an
{\sc is-a} hierarchy.  To illustrate how our methods can be used to
measure inter-annotator agreement, we show how to compute the kappa
coefficient over hierarchical tag sets.
\end{abstract}
\end{opening}

\section{Introduction}

Objective evaluation has been central in advancing our understanding
of the best ways to engineer natural language processing systems.  A
major challenge of objective evaluation is to design fair and
informative evaluation metrics, and algorithms to compute those
metrics.  When the task involves any kind of tagging (or
``labeling''), the most common performance criterion is simply ``exact
match,'' i.e. exactly matching the right answer scores a point, and no
other answer scores any points.  This measure is sometimes adjusted
for the expected frequency of matches occuring by chance
\cite{carletta1996}.  Resnik and Yarowsky
\shortcite{ry1997:siglex,ry1998:nle}, henceforth R\&Y, have argued
that the exact match criterion is inadequate for evaluating word sense
disambiguation (WSD) systems.

R\&Y proposed a generalization capable of assigning partial credit,
thus enabling more informative comparisons on a finer scale.  In this
article, we present three further generalizations.  First, we show how
to evaluate non-probabilistic assignments of multiple tags.  Second,
we show how to accommodate a common property of manually constructed
``gold standards'' that are typically used for objective evaluation,
namely that there is often more than one correct answer.  Third, we
show how to measure performance when the set of possible tags is
tree-structured in an {\sc is-a} hierarchy.  To illustrate how our
methods can be applied to the comparison of human taggers, we show how
to compute the kappa coefficient \cite{kappa} over hierarchical tag
sets.

Our methods depend on the tree structure of the tag hierarchy, but not
on the nature of the nodes in it.  For example, although these
generalizations were motivated by the {\sc senseval} exercise (Palmer
and Kilgarriff, this issue), the mathematics applies just as well to
any tagging task that might involve hierarchical tag sets, such as
part-of-speech tagging or semantic tagging \cite{muc7}.  With respect
to word sense disambiguation in particular, questions of whether
part-of-speech and other syntactic distinctions should be part of the
sense inventory are orthogonal to the issues addressed here.

\section{Previous Work}

Work on tagging tasks such as part-of-speech tagging and word sense
disambiguation has traditionally been evaluated using the exact match
criterion, which simply computes the percentage of test instances for
which exactly the correct answer is obtained.  R\&Y noted that, even
if a system fails to uniquely identify the correct tag, it may
nonetheless be doing a good job of narrowing down the possibilities.
To illustrate the myopia of the exact match criterion, R\&Y used the
hypothetical example in Table~\ref{tbl:senses}.  Some of the systems
in the table are clearly better than others, but all would get zero
credit under the exact match criterion.
\begin{table}
% \begin{footnotesize}
\begin{center}
\caption{Hypothetical output of four WSD systems on a test instance,
where the correct sense is (2).  The exact match criterion would
assign zero credit to all four systems.  Source: \protect\cite{ry1997:siglex} }
\label{tbl:senses}
\begin{tabular*}{\maxfloatwidth}{lrrrr}
\hline
        		  			&
        		  			\multicolumn{4}{c}{WSD
        		  			System} \\
\multicolumn{1}{c}{sense of {\em interest} (in English)} 	& 1 & 2 & 3 & 4 \\
\hline
(1) monetary (e.g. on a loan)   		& .47 & .85 & .28 & 1.00 \\
(2) stake or share {\Large $\Leftarrow$ {\bf correct}} 	& .42 & .05 & .24 &  .00 \\
(3) benefit/advantage/sake      		& .06 & .05 & .24 &  .00 \\
(4) intellectual curiosity      		& .05 & .05 & .24 &  .00 \\
\hline
\end{tabular*}
\end{center}
% \end{footnotesize}
\end{table}

R\&Y proposed the following measure, among others, as a more
discriminating alternative:
\begin{eqnarray}
\label{RYeqn}
Score({\cal A}) = \Pr_{{\cal A}}(c | w,\mbox{context}(w)),
\end{eqnarray}
In words, the score for system ${\cal A}$ on test instance $w$ is the
probability assigned by the system to the correct sense $c$ given $w$
in its context.  In the example in Table~\ref{tbl:senses}, System~1
would get a score of 0.42 and System~4 would score zero.

\section{New Generalizations}
The generalizations below start with R\&Y's premise that, given a
probability distribution over tags and a single known correct
tag, the algorithm's score should be the probability that the
algorithm assigns to the correct tag.

% \begin{itemize}
\subsection{Non-probabilistic Algorithms}
% \item[\bf Non-probabilistic Algorithms] 
Algorithms that output multiple tags but do not assign probabilities
should be treated as assigning uniform probabilities over the tags
that they output.  For example, an algorithm that considers tags A and
B as possible, but eliminates tags C, D and E for a word with 5 tags
in the reference inventory should be viewed as assigning probabilities
of .5 each to A and B, and probability 0 to each of C, D, and E.
Under this policy, algorithms that deterministically select a single
tag are viewed as assigning 100\% of the probability mass to that one
tag, like System~4 in Table~\ref{tbl:senses}.  These algorithms would
get the same score from Equation~\ref{RYeqn} as from the exact match
criterion.

\subsection{Multiple Correct Tags}
% \item[\bf Multiple Correct Tags] 
Given multiple correct tags for a
given word token, the algorithm's score should be the sum of {\em
all\/} probabilities that it assigns to {\em any\/} of the correct
tags; that is, multiple tags are interpreted disjunctively.  This is
consistent with instructions provided to the {\sc senseval}
annotators: ``In general, use disjunction ... where you are unsure
which tag to apply'' \cite{krishnamurthy1998:senseval}.  In symbols,
we build on Equation~\ref{RYeqn}:
\begin{eqnarray}
\label{multi_eqn}
Score({\cal A}) = \sum_{t=1}^{C} \Pr_{{\cal A}}(c_t | w,\mbox{context}(w)),
\end{eqnarray}
where $t$ ranges over the $C$ correct tags.  Even if it is impossible
to know for certain whether annotators intended a multi-tag annotation
as disjunctive or conjunctive, the disjunctive interpretation gives
algorithms the benefit of the doubt.  

\subsection{Tree-structured Tag Sets}
% \item[\bf Tree-structured Tag Sets] 
The same scoring criterion can be
used for structured tag sets as for unstructured ones: What is the
probability that the algorithm assigns to any of the correct tags?
The complication for structured tag sets is that it is not obvious how
to compare tags that are in a parent-child relationship.  The
probabilistic evaluation of taggers can be extended to handle
tree-structured tag sets, such as {\sc hector} \cite{atkins1993}, if
the structure is interpreted as an {\sc is-a} hierarchy.  For example,
if word sense A.2 is a sub-sense of word sense A, then any word token
of sense A.2 {\em also\/} {\sc is-a} token of sense A.  

Under this interpretation, the problem can be solved by defining two
kinds of probability distributions:

% \begin{footnotesize}
\begin{enumerate}
\item $\Pr(\mbox{occurrence of parent tag} |
           \mbox{occurrence of child tag})$ 
\item $\Pr(\mbox{occurrence of child tag} | 
           \mbox{occurrence of parent tag})$.  
\end{enumerate}
% \end{footnotesize}

\noindent In a tree-structured {\sc is-a} hierarchy $\Pr(\mbox{parent}
| \mbox{child}) = 1$, so the first one is easy.  The second one is
harder, unfortunately; in general, these ("downward") probabilities
are unknown.  Given a sufficiently large training corpus, the downward
probabilities can be estimated empirically.  However, in cases of very
sparse training data, as in {\sc senseval}, such estimates are likely
to be unreliable, and may undermine the validity of experiments based
on them.  In the absence of reliable prior knowledge about tag
distributions over various \mbox{tag-tree} branches, we appeal to the
maximum entropy principle, which dictates that we assume a uniform
distribution of sub-tags for each tag.  This assumption is not as bad
as it may seem.  It will be false in most individual cases, but if
we compare tagging algorithms by averaging performance over many
different word types, most of the biases should come out in the wash.

Now, how do we use these conditional probabilities for scoring?  The
key is to treat each non-leaf tag as under-specified.  For example, if
sense A has just the two subsenses A.1 and A.2, then tagging a word
with sense A is equivalent to giving it a probability of one half of
being sense A.1 and one half of being sense A.2, given our assumption
of uniform downward probabilities.  This interpretation applies both
to the tags in the output of tagging algorithms and to the manual
(correct, reference) annotations.
% \end{itemize}

\section{Example}

\label{egsection}

Suppose our sense inventory for a given word is as shown in Figure~1.
\begin{figure}[htb]
\centerline{\psfig{figure=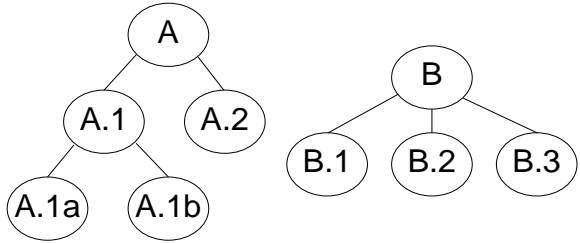,width=3in}}
\caption{Example tag inventory. \label{egtree}}
\end{figure}
\newline
Under the assumption of uniform downward probabilities, we start by
deducing that \( \Pr(A.1 | A) = .5, \Pr(A.1a | A.1) = .5,\) \mbox{(so
\( \Pr(A.1a | A) = .25\) ),} \(\Pr(B.2 | B) = \frac{1}{3}, \) and so
on. If any of these conditional probabilities is reversed, its value
is always 1.  For example, $\Pr(A | A.1a) = 1$.  Next, these
probabilities are applied in computing Equation~\ref{multi_eqn}, as
illustrated in Table~\ref{tbl:examples}.

\begin{table}
\begin{center}
\caption{Examples of the scoring scheme, for the tag inventory in Figure~1.}
\label{tbl:examples}
\begin{footnotesize}
\begin{tabular}{ccc}
\hline
Manual Annotation	& Algorithm's Output	& Score \\
\hline
 B			& A		& 0 \\
 A			& A		& 1 \\
 A			& A.1		& 1 \\
 A			& A.1b		& 1 \\
 A.1			& A		& .5 \\
 A.1 and A.2		& A		& .5 + .5 = 1 \\
 A.1a			& A		& .25 \\
 A.1a and B.2		& B		& $\Pr(B.2 | B) = \frac{1}{3}$ \\
 A.1a and B.2		& A.1	        & .5 \\
 A.1a and B.2		& A.1 and B.2   & $.5 \times .5 + .5 \times 1 = .75$ \\
 A.1a and B.2		& A.1 and B & $.5 \times .5 + .5 \times .333 = .41666$ \\
\hline
\end{tabular}
\end{footnotesize}
\end{center}
\end{table}

\section{Inter-Annotator Agreement Given Hierarchical Tag Sets}

Gold standard annotations are often validated by measurements of
inter-annotator agreement.  The computation of any statistic that may
be used for this purpose necessarily involves comparing tags to see
whether they are the same.  Again, the question arises as to how to
compare tags that are in a parent-child relationship.  We propose the
same answer as before: Treat non-leaf tags as underspecified.

To compute agreement statistics under this proposal, every
non-leaf tag in each annotation is recursively distributed over
its children, using uniform downward probabilities.  The resulting
annotations involve only the most specific possible tags, which can
never be in a parent-child relationship.  Agreement statistics can
then be computed as usual, taking into account the probabilities
distributed to each tag.  

One of the most common measures of pairwise inter-annotator agreement
is the kappa coefficient \cite{kappa}:
\begin{equation}
K = \frac{\Pr(A) - \Pr(E)}{1 - \Pr(E)}
\end{equation}
where $\Pr(A)$ is the proportion of times that the annotators agree
and $\Pr(E)$ is the probability of agreement by chance.  Once the
annotations are distributed over the leaves $L$ of the tag inventory,
these quantities are easy to compute.  Given a set of test instances
$T$,
\begin{equation}
\Pr(A) = \frac{1}{|T|} \sum_{t \in T} \sum_{l \in L} \Pr(l | \mbox{annotation}_1(t)) \cdot \Pr(l | \mbox{annotation}_2(t))
\end{equation}
\begin{equation}
\Pr(E) = \sum_{l \in L} \Pr(l)^2
\end{equation}
Computing these probabilities over just the leaves of the tag
inventory ensures that the importance of non-leaf tags is not
inflated by double-counting.

\section{Conclusion}

We have presented three generalizations of standard evaluation methods
for tagging tasks.  Our methods are based on the principle of maximum
entropy, which minimizes potential evaluation bias.  As with the R\&Y
generalization in Equation~\ref{RYeqn}, and the exact match criterion
before it, our methods produce scores that can be justifiably
interpreted as probabilities.  Therefore, decision processes can
combine these scores with other probabilities in a maximally
informative way by using the axioms of probability theory.

Our generalizations make few assumptions, but even these few
assumptions lead to some limitations on the applicability of our
proposal.  First, although we are not aware of any algorithms that
were designed to behave this way, our methods are not applicable to
algorithms that {\em conjunctively\/} assign more than one tag per test
instance.  A potentially more serious limitation is our interpretation
of tree-structured tag sets as \mbox{{\sc is-a}} hierarchies.  There
has been considerable debate, for example, about whether this
interpretation is valid for such well-known tag sets as {\sc hector}
and WordNet.

This work can be extended in a number of ways.  For example, it would
not be difficult to generalize our methods from trees to hierarchies
with multiple inheritance, such as WordNet \cite{fellbaum1998}.

%%%%%%%%%%%%%%%%%%%%%%%%%%%%%%%%%%%%%%%%%%%%%%%%%%%%%%%%%%%%%%%%
% \bibliographystyle{klunamed}
% \bibliography{resnik,senses,bible,bibliography,general,learning,distrib,nlstat,ibm_master,gisting,ir}
%%%%%%%%%%%%%%%%%%%%%%%%%%%%%%%%%%%%%%%%%%%%%%%%%%%%%%%%%%%%%%%%

\end{article}
\end{document}